\documentclass[runningheads]{llncs}
\usepackage[T1]{fontenc}
\usepackage{graphicx}
\usepackage{xspace}
\usepackage{enumitem}
\usepackage{subcaption}

\begin{document}
\newcommand{\sysname}{\emph{PTStore}\xspace}
\title{\sysname (Prefix Tensor Store): Distributed Prefix Caching and Replication for High Throughput Inference Serving}
\titlerunning{PTStore}

\author{Meghana Maghyastha\inst{1}\and
Robert Underwood\inst{2}\and
Randal Burns\inst{1}\and
Bogdan Nicolae\inst{2}\thanks{Corresponding author.}}
\authorrunning{M. Maghyastha et al.}

\institute{Johns Hopkins University, USA --
\email{\{meghana,randal\}@cs.jhu.edu}\and
Argonne National Laboraotory, USA --
\email{\{runderwood,bnicolae\}@anl.gov}}

\maketitle              
\begin{abstract}
Inspired by the design of client caching in Content Delivery Networks (CDNs), PTStore distributes and replicates popular tensors that form reusable KV cache prefixes, which are the main technique used by state of art approaches to accelerate inferences. This reduces the latency of accessing the KV cache and alleviates load imbalance caused by a disproportionately large number of requests on servers containing popular tensors. Furthermore, thanks to decentralization, PTStore allows the expansion of the size of the KV cache for LLM inference by orders of magnitude. As a result, PTStore can execute inferences on long passage Q\&A datasets 5-6 times more efficiently than current baselines, which do not aggregate memory across different nodes and GPUs and therefore require regenerating the KV cache.
\end{abstract}

\section{Introduction}
\label{sec:intro}
Large Language Models (LLMs) and foundation models (FMs) have revolutionized multiple domains by enabling the understanding and generation of human-like text with remarkable fluency and coherence. These models, trained on vast amounts of data, can perform a wide range of tasks: literature search~\cite{tilwani2024reasons}, knowledge distillation~\cite{gou2021knowledge}, and complex reasoning~\cite{bubeck2023sparks}. They enable researchers to navigate complex scientific problems more efficiently.

LLM/FM training and fine-tuning pose a significant challenge by requiring tens of thousands of GPUs for several months at a time. Until recently, these workloads were dominating HPC data centers. However, inference workloads have become the dominant workloads nowadays. With an exploding number of LLM users, serving inference requests both efficiently and at scale is emerging as an even bigger challenge. For example, at Meta, inferences are the largest AI workload running on their HPC data centers, accounting for 65\% of the energy consumption, compared to 35\% consumption for pre-training \cite{Castro2024AIenergy}.

LLM inferences are challenges because they involve two phases that use resources in different ways: prefill and decode. During the prefill, the input prompt is processed in parallel, which is compute-intensive and effectively utilizes the GPU's processing power to generate the first token. Then, successive tokens
are generated sequentially in auto-regressive fashion in decode steps. To avoid recomputations associated with the attention mechanism (which captures correlations between pairs of tokens, most of which do not change between decode steps),
a KV cache~\cite{pope2023efficiently} is often employed to store the K and V vectors (reusable intermediate attention results) of previously computed tokens. The KV cache typically resides in the spare GPU memory not occupied by the model parameters. The same principle can be applied across different inference requests that share the same prefix. In this case, the K and V vectors of the longest common prefix can be reused to accelerate the prefill by avoiding redundant attention computations.

\textbf{Limitations of state of art.}
Modern inference runtimes such as vLLM~\cite{vLLM-kwonEfficientMemoryManagement2023} employ sophisticated
KV cache techniques that allow non-contiguous allocation of blocks of GPU memory where the KV results of multiple requests that are batched together and processed in parallel can be stored. To enable prefix caching, a simple approach is to keep KV blocks in the KV cache for as long as possible, subject to an eviction policy
(e.g., apply LRU to KV blocks of requests that finished). Then, if a new request
arrives that shares a common prefix with a previous request (either finished and not evicted yet or in progress), the blocks corresponding to the longest common
prefix can simply be reused (and marked as such by increasing a reference counter). More sophisticated extensions to this technique can be adopted where KV blocks are proactively flushed from GPUs to the host memory (shared by all GPUs on the same compute node), which extends the KV cache capacity of the individual GPUs and
allows the reuse of prefixes across requests served by different GPUs, at
the expense of slower access. Such an approach is implemented by runtimes such as LMCache~\cite{liu2025lmcacheefficientkvcache}. However, at scale, inference requests are served by a large number of GPUs distributed over a large number of compute nodes. In this case, prefixes corresponding to requests served by the GPUs on some nodes can be reused by the GPUs on other nodes. This aspect is insufficiently addressed by state of art,
leading to missed opportunities.

\textbf{Contributions.} In this paper, we focus on the challenge of how to enable scalable reuse of KV cache prefixes across a large number of GPUs distributed across many compute nodes. The key idea we adopt is to distribute and store the KV cache prefixes at fine granularity (tensor-level) using an incremental technique, which allows the prefixes to grow redundancy-free over time in divergent directions, similar to a trie. At the same time, we introduce consolidated metadata
techniques that allow efficient one-shot prefix queries without the need to walk
a distributed trie. We complement these ideas with an innovative prefix replication
strategy that improves the locality of accesses to popular ``hot'' prefixes at the expense of redundancy that is reconciled with the scarce memory capacity available
for the KV cache. We summarize our contributions as follows:

\begin{enumerate}[topsep=0pt,noitemsep,leftmargin=12pt]
\item We present a series of high-level design principles for a distributed repository that integrates the above ideas(\S~\ref{sec:approach}).
\item We detail \sysname, a research
prototype that illustrates the design principles through
a practical implementation. \sysname enables high I/O throughput under concurrency and integrates seamlessly with other LLM runtimes (\S~\ref{sec:impl}).
\item We run extensive experiments at scale highlighting the significant reduction in I/O overheads and overall end-to-end runtime compared with other state-of-the-art approaches (\S~\ref{sec:exp}).
\end{enumerate}

\section{Background}
\label{sec:background}

\vspace{-20pt}
\begin{figure*}
    \centering
    \includegraphics[width=0.85\linewidth]{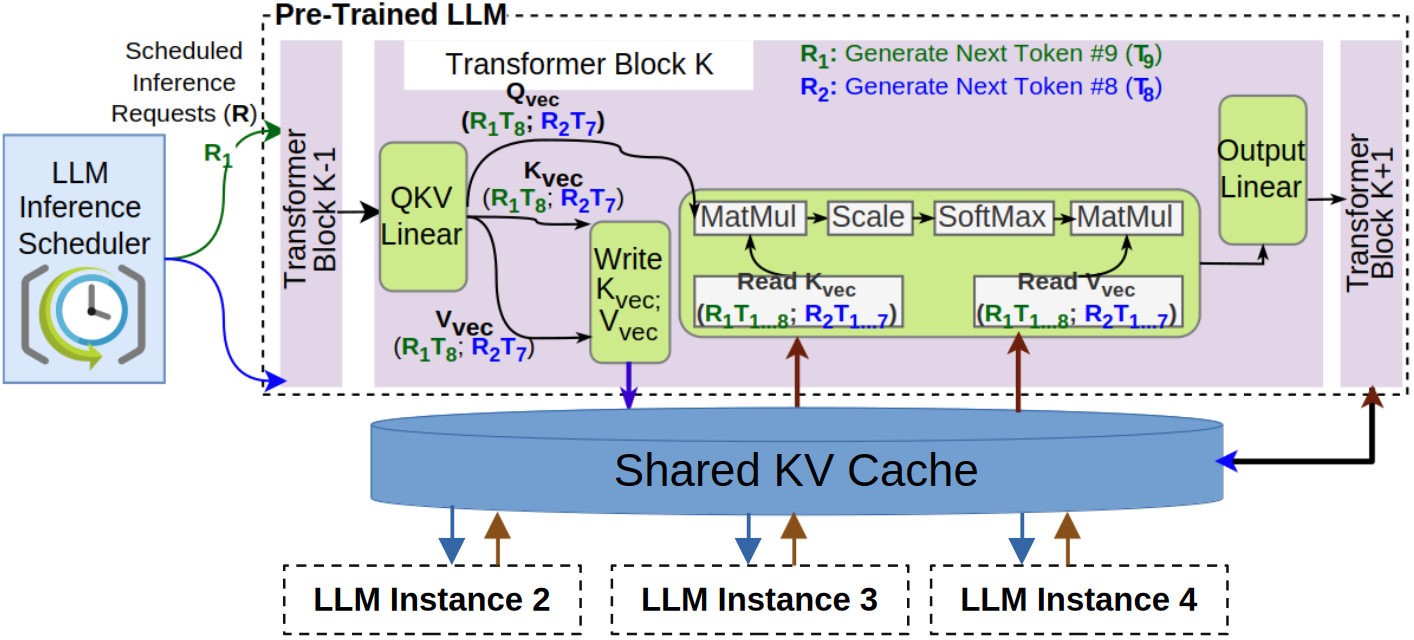}
    \caption{Overview of shared KV caching for LLM inference. Multiple transformer blocks access the KV cache to incrementally store and reuse KV results. The KV cache may be shared by multiple LLM instances to exploit shared prefixes.}
    \label{fig:inferenceoverview}
\end{figure*}

Just like in the case of regular deep learning models, LLM inferences are based on \emph{forward passes} that take a prompt (a sequence of tokens) as input and generate a reply as output. The reply is the most likely sequence of tokens that continues the prompt, similar in scope to sequence-to-sequence models. Unlike regular deep learning models, the reply is constructed iteratively, one token at a time. This happens in two phases. First, the \emph{prefill} phase generates the first output token. Then, the output token is appended to the prompt, and the next output token is generated in the \emph{decode} step. The decoding step is repeated until a maximum number of output tokens is reached or a special termination token (\texttt{<EOS>}) is generated. The initial prefill and the successive decode steps run each in a separate forward pass.

A key component of typical transformer architectures is the attention layers, often organized in multiple heads~\cite{Attention-NIPS17}. This is illustrated in Figure~\ref{fig:inferenceoverview}. Attention layers capture positional correlations between pairs of tokens using multiplications of multi-dimensional matrices (denoted Q, K, and V), which can take advantage of the massive parallelism offered by GPUs. After the prefill phase, the prompt has only changed by one appended token. Thus, during the decode phase, it is enough to compute and incrementally store only the intermediate K and V vectors of the new token (for each head and each layer), while reusing the cached K and V vectors for all other tokens.

As more inference requests keep being served over time, it is often the case that different inference requests, potentially belonging to different users, share the same prefix (e.g., queries about the same text or conversations that build on previous questions and answers)~\cite{juravsky2024hydragenhighthroughputllminference}. In this case,
reusing the K and V vectors corresponding to the longest common prefix between the prompt of a previous request and the prompt of a new request accelerates
the prefill similarly to how reusing the K and V vectors accelerates the decode steps of the same request. Thus, prefix caching is increasingly being employed as a core optimization to improve the performance and scalability of inference serving.

\section{Related Work}
\label{sec:related}
Inference frameworks such as vLLM \cite{vLLM-kwonEfficientMemoryManagement2023} and DeepSpeed-MII \cite{microsoftDeepSpeedFastGenIntroducingMixtral} dramatically decrease the time to first token (TTFT) and overall inference throughput thanks to caching of intermediate results (K V vectors) of attention layer computations. However, these techniques mostly focus on GPU memory and lack support for aggregating distributed memory tiers.  Multi-level caching using disaggregated resource management is an aspect leveraged by systems like Splitwise~\cite{patel2024splitwiseefficientgenerativellm}, DistServe~\cite{zhong2024distservedisaggregatingprefilldecoding}, and TetriInfer~\cite{hu2024inferenceinterferencedisaggregatellm}. The latter separates the prefill and decode phases, allowing their scheduling and batching on different compute nodes. SwapAdvisor~\cite{SwapAdvisor} uses genetic algorithms to control memory allocations and swap decisions. vDNN~\cite{vdnn} employs offloading and prefetching. TSPLIT~\cite{9835178} uses tensor splitting to enable fine-grain control of the KV cache. vLLM~\cite{vLLM-kwonEfficientMemoryManagement2023} uses either recompute or swap, implementing an all-or-nothing eviction policy configurable by the user. On the other hand, STR~\cite{10098636} can dynamically combine both techniques. Most of these approaches are complementary to our own work, focusing on KV caching optimizations for independent queries. On the other hand, our approach focuses on \emph{distributed caching} of shared prefixes between \emph{different} queries.

Building on the concept of shared prefix management, systems like LMCache \cite{liu2025lmcacheefficientkvcache}, EvoStore~\cite{underwood2024evostore}, MoonCake~\cite{Mooncake25} and SGLang~\cite{SGLang24} can be used to extend KV caching beyond the boundaries of a single request. SGLang introduces RadixAttention, which treats the KV cache as a radix tree, allowing for automatic and efficient sharing of overlapping prefixes between different prompts. While RadixAttention excels at intra-node cache reuse,
EvoStore specializes in capturing evolving groups of tensors in a distributed fashion. While originally explored for network architecture search, the same prefix sharing capabilities can be leveraged in the context of inference serving. Mooncake introduces a KVCache-centric disaggregated architecture that utilizes a tiered storage hierarchy, spanning GPU memory, local DRAM, and remote SSDs. Similarly, LMCache enables the sharing of KV caches across different conversation sessions and even separate serving engine instances, significantly reducing redundant computations for popular system prompts or long-context documents. However, these systems face critical bottlenecks either in metadata synchronization (e.g. scaling
RadixAttention beyond a single node is difficult) or excessive remote transfers
(even if performed using RDMA), which introduce high I/O overheads that offset
the benefits of accelerating compute-heavy prefilling.

\section{\sysname: Distributed Replicated Prefix Tensor Store}
\label{sec:approach}

\begin{figure}[ht]
    \centering
    \includegraphics[width=0.8\columnwidth]{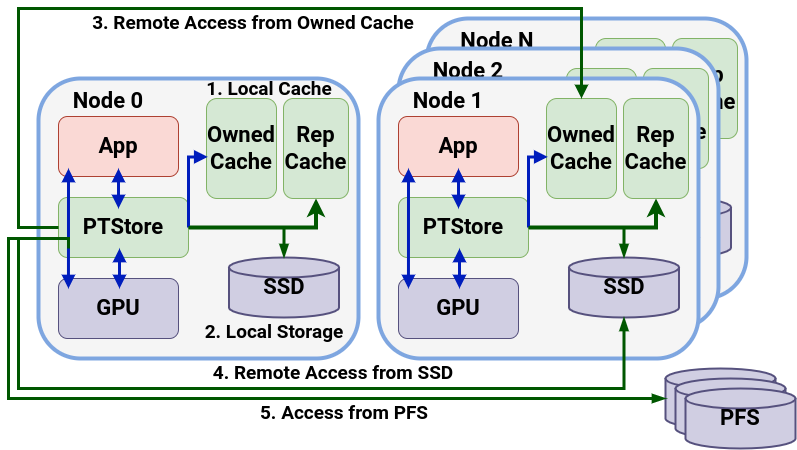}
      \caption{Distributed architecture of \sysname. Each compute node deploys one server that aggregates the node's host memory and node-local SSDs, serving all GPU clients (both local and remote). Our contributions are highlighted in green.}
    \label{fig:sysdesign}
\end{figure}

In this section, we introduce our contribution: \sysname, a scalable repository for fine-grain, multi-level tensor storage and prefix caching. The architecture of \sysname is illustrated in Figure~\ref{fig:sysdesign}.

In summary, we aggregate the memory tiers of a large number of compute nodes available in a data center or HPC machine by running a distributed runtime that deploys a server on each compute node. Each server is co-located with the application processes (clients) running on the same compute node. Store operations save incremental differences (a group of tensors) between a new object (KV intermediate results) and a previously stored object. To minimize the size of the incremental difference, save operations identify the longest common prefix
between this object and any other previously stored object. The server that executes the store operation becomes the owner of the increment, which is stored in the owned cache. As new objects are stored, increments form increasingly longer prefixes that can be shared across all objects. To speed up load operations, each server caches non-owned increments that are part of popular prefixes in the replication cache. We leverage a distributed hierarchical caching strategy that is responsible for replicating and retaining the most popular prefixes on the fastest memory tiers. For the rest of this section, we focus on the general design principles and zoom in on several implementation details.

\subsection{Design Principles}

\textbf{Co-optimized compact metadata and fast prefix queries:}
For each stored object, we need to retain its composition in terms of tensors.
To avoid multiple levels of indirection or expensive reconstruction of the composition by iterating over all overlapping prefixes, we propose a single flat metadata structure that holds a list of unique tensor IDs. Thus, each object
will inherit the unique IDs of the longest common prefix shared with other objects
and will append its own unique IDs representing the incremental difference.
Each unique ID not only identifies the tensor, but also the server that stores
it. Therefore, a load operation simply needs to iterate over the unique IDs
of an object and check if the tensors already exist locally in the replication cache. If they don't, then they will be fetched remotely from the servers that
own those tensors, since the tensors are guaranteed to exist in the owner cache
hierarchy. A store operation needs to first find out the longest common prefix
among all previously stored objects. To this end, we implement a two-level parallel reduction among the servers. First, since the payload contains only compact tensor IDs (a few bytes per prefix), the LCP query remains bandwidth-light as it is distributed at large scale across many servers. Second, since the metadata of each object is self-contained and does not depend on the metadata of other objects, each server can run efficient parallel matches against its locally owned cache hierarchy. Then, using the longest common prefix, the incremental difference
is stored randomly on one of the servers, which enables lightweight
load balancing without synchronization (both in terms of cache hierarchy utilization
and metadata).

\textbf{Fine-grain multi-level caching of prefixes with configurable eviction policy:}
Each server manages multiple memory tiers available on the compute node where it is deployed: host memory, SSDs, and external storage (e.g., parallel file systems or cloud storage). The memory tiers are split between the owned cache and the
replication cache. The last level of the owned cache guarantees persistence
for the stored tensors. Conversely, tensors on the replication cache can be evicted
from any level. When issuing load/store operations, the application supplies pointers to the GPU memory where the tensors are located that make up a new object to be loaded/stored. It is the responsibility of \sysname to interact with the
rest of the storage hierarchy in transparent fashion in order to implement caching, persistence and replication. To minimize I/O blocking and enable overlapping
with computations, data transfers happen asynchronously and are governed by
a configurable eviction policy.

\textbf{Replication of popular prefixes to improve access locality:}
As an object begins to be scattered across multiple servers, load operations need to contact multiple servers who own the tensors. Even if such operations can be executed in parallel, their latency can be significantly higher than local host memory accesses. As a consequence, it is important to replicate ``hot'' tensors locally, which is the responsibility of the replication cache. However, the
owner cache and the replication cache compete for the same local storage
hierarchy (host memory, SSDs). How to split the host memory between caching owned tensors vs. replicated tensors is subject to a trade-off: Faster retrieval of hot tensors cached locally may be offset by a higher penalty of retrieving cold tensors (because these tensors are more likely to be on slower tiers on remote servers due to less space available to cache-owned tensors). Our solution allows a configurable threshold, which can be fine-tuned together with the eviction policy.

\textbf{Access pattern-aware eviction policy:}
The unique overlapping structure of prefixes lends itself to various optimizations. First, tensors at the front of a prefix are more likely to be accessed as part of a longest common prefix. However, popular eviction criteria such as LRU (Least Recently Used) often fail to capture structural aspects like this. We empirically observe that frequency-based eviction criteria outperform LRU for this reason. Furthermore, when the objects and prefixes are made of tensors that are of variable size, it is important to avoid the situation when many small, frequently used tensors limit the opportunity to cache larger tensors that are more expensive to fetch from a remote memory tier. As a consequence, to account for the size vs. frequency trade-off, we adapt GDSF~\cite{GDSF97,gdsfweb}, an eviction policy originally designed for web services, for use with shared prefixes. Note that the combination of replicated and owned tensors further complicates eviction decisions. Replicated tensors can be safely dropped at the expense of triggering remote I/O to other compute nodes. Owned tensors need to be flushed to slower storage tiers. As a consequence, we put a lower and upper limit on the capacity reserved for replicated tensors, and prioritize discarding the replicated tensors as long as the lower limit is not reached, before evicting owned tensors to slower memory tiers.

\textbf{RDMA-aware consolidation in host memory:}
To avoid excessive scattering of the tensors across multiple servers, the increments appended to the longest common prefixes are consolidated as a single contiguous region on a single server (which becomes the owner of the tensors).
When the corresponding server (or servers) are not co-located on the same compute node as the client, then load/store operations need to bring the tensors in host memory first (if they are not already present there), from where they
are read or written into remotely using RDMA. To this end, we leverage bulk RDMA operations that enable a zero-copy approach:  each tensor is assigned an (offset, size) segment, and all scattered segments are accessed in parallel in a single
RDMA RPC request. This enables faster access to scattered tensors compared with copying the tensors into a single contiguous region that is then transferred as a single segment. As a further optimization, we cache the RDMA segments together with the tensors to avoid expensive RDMA setup operations for subsequent load operations.

\section{Implementation}
\label{sec:impl}

We designed and built a software prototype based on the principles in Section~\ref{sec:approach}. Specifically, we follow a client-server design where each client interacts with several servers in parallel. The clients utilize a low-level library that the applications link with. It exposes a C++ low-level API to issue the longest common prefix (LCP) queries (which transparently broadcast and reduce the results) and to read/write subsets of tensors. The client is responsible for interpreting the metadata, which indicates that the server is hosting the relevant prefixes as a composition of tensors. Based on the metadata, it loads and stores the prefixes using optimized RDMA communications that progress with multiple servers in parallel. To this end, we leveraged an optimized HPC-oriented remote procedure calls (RPCs) strategy based on bulk RDMA operations, as provided by the Mochi~\cite{rossMochiComposingData2020} collection of composable building blocks. Specifically, we use \textbf{Thallium}, which is a C++ wrapper on top of Mercury and Argobots. The servers use an extensible key-value store abstraction that can be used to group and store tensors at fine granularity, either in memory or to serialize them persistently into files on SSDs or
parallel file systems. The key-value store abstraction is tightly coupled with our memory allocator to reduce latency and improve bandwidth of RDMA by avoiding unnecessary registration overheads. It is also general enough to represent prefixes as any combination of tensors of variable sizes.

To interface with \textbf{vLLM}~\cite{vLLM-kwonEfficientMemoryManagement2023},
we modified the part of vLLM that handles prefix sharing by replacing the original code that issues accesses to the host memory with corresponding RPC invocations to \sysname. To this end, we implemented a Python interface for \sysname that relies on the low-level C++ client API to expose more concise, Pythonic primitives at a higher level. These primitives handle both LCP queries and load/store operations in a user-friendly fashion without sacrificing performance. Specifically, the bridging between C++ and Python is based on \textbf{nanobind}. Nanobind was chosen for its implementation of the dlpack protocol, which enables efficient and interoperable access to GPU and CPU memory from various Python data frameworks, and low-overhead binding compared to alternatives like Pybind11.

\section{Experimental evaluation}
\label{sec:exp}

\subsection{Methodology}
\label{sec:method}

\paragraph*{\bf Experimental setup}
\label{sec:setup}
We conduct our experiments on ALCF’s Polaris HPC testbed. It consists of 560 nodes, each equipped with 512 GB of DDR4 memory (aggregated from four NUMA domains), a 32-core AMD Zen 3 (Milan) (64 threads), two 1.6 TB SSDs (2 GB/s), and four Nvidia A100 GPUs aggregating to a total of 160 GB HBM memory. On each node, the four A100 GPUs are connected with four NVLinks and with the host memory through a PCIe Gen 4 interface. The peak unidirectional Device-to-Device (D2D), nd pinned Device-to-Host (D2H) (and vice versa) bandwidths on each GPU are 85 GB/s and 25 GB/s, respectively. There is a one-to-one mapping between the GPU and the NUMA domains, therefore, concurrent device-to-host access by multiple GPUs does not create contention on the PCIe interface. The compute nodes are interconnected with a dual Slingshot 10 network fabric. Persistent storage is provided by a Lustre parallel file system, composed of 160 Object Storage Targets (OSTs) and 40 Metadata targets, with an aggregated bandwidth of 650 GB/s.

\paragraph{\bf Workloads: LLM Inferences for Extractive Tasks.}
We use two question-answer workloads that feature long inference queries. These workloads focus on extractive tasks~\cite{abney2000answer} in which the LLM needs to identify where in a given document the answer to a specific question about the content of that document is. The format of the inference query is a long prompt and short answer, which emphasizes the prefill stage. For each text, there are multiple different questions, resulting in large prefixes of the KV cache that can be reused across different queries. Details of the datasets used for the two workloads are as follows:
\begin{enumerate}[leftmargin=12pt]
      \item \textbf{WikiQA Dataset:} The Microsoft WikiQA dataset~\cite{yang2015wikiqa} consists of 3000 questions in which each question is associated with a Wikipedia page that contains the answer. The page lengths range from $2,000$ to $40,000$ tokens. We use this dataset for scaling experiments because the pages used as context in the prefill stage are long, which creates challenging long prefixes.

     \item \textbf{SQUAD Dataset:} This dataset focuses on reading comprehension and includes 100,000 questions. Similar to the previous dataset, each question is associated with a passage that contains an answer. Passage lengths range from 150 words to 4000 words. We use SQUAD for experiments that vary sequence length, because of the large number of questions produced in many sequences at each length, which creates a rich diversity of prefixes.
\end{enumerate}

\paragraph{\bf Inference Setup.}
The LLM used to perform the inferences is the \emph{Mistral-7B-instruct-V2} model. We replicate this model on each GPU (it is small enough to fit in the memory of an A100 GPU, while leaving at least 15\% of GPU memory for KV caching). The runtime used to perform the inferences is vLLM (v0.6). Each worker (one per GPU) performs 1500 inference queries (a pair consisting of the document content and a question about the document). For \sysname, we cap the replication cache at 50\% of each server's host memory budget, with the remaining capacity reserved for owned tensors; a more detailed sensitivity sweep over this ratio is left to future work.

\paragraph{\bf Metrics.} To generate inference queries, we sample documents and corresponding questions using a power law with parameter $\alpha=6$~\cite{newman2005power}, chosen to mirror the heavy-tailed prefix-popularity skew observed in production LLM serving traces and to stress the replication cache by concentrating reuse on a small subset of prefixes. We measure the average time to first token (TTFT), since extractive tasks produce a single-token answer (the position in the document where the answer can be found).

\subsection{Compared approaches}

Throughout our evaluations, we compare the following four approaches, each of which progressively adds more prefix reuse capabilities.

\paragraph{\bf vLLM Default KV Cache Block Manager (vLLM Vanilla).}
We use the KV cache block manager of vLLM~\cite{vLLM-kwonEfficientMemoryManagement2023} as a baseline for comparison. As mentioned in \S~\ref{sec:impl}, this implementation primarily leverages the spare GPU memory as a KV cache. We activate swapping of KV cache blocks to the host memory to avoid
recomputations in case some requests need to be preempted and resumed later. By default, vLLM does not index and look for shared prefixes that can be reused.
Therefore, despite avoiding recomputations for the same request thanks to swapping,
recomputations are triggered across different requests due to lack of shared prefix awareness. We set the vLLM KV cache block size to 1024 tokens.

\paragraph{\bf vLLM Prefix-Aware KV Cache Block Manager (vLLM Prefix).}
The KV cache block manager of vLLM can make use of shared prefixes if it is  explicitly configured to do so. In this case, we use the same configuration
as above (vLLM Vanilla), but we allow shared prefixes to be reused. In this
case, each GPU manages its own KV cache and can independently reuse local
prefixes. Furthermore, prefixes can be offloaded to a common host memory pool
and reused across the node-local GPUs through extensions such as
LMCache~\cite{liu2025lmcacheefficientkvcache}, but this is a limited capability that does not allow sharing prefixes across GPUs on different nodes. In terms of attention implementation, we use the flash kernel modified to be prefix-aware for vLLM~\cite{flashinfer}.

\paragraph{\bf EvoStore backend for vLLM KV Cache Block Manager (EvoStore).}
We add distributed KV cache prefix awareness across different nodes on top of \emph{EvoStore}~\cite{underwood2024evostore}, a state-of-the-art tensor store that is based on the idea of aggregating the spare DRAM of the compute nodes to distribute the tensors. It employs incremental storage techniques and RDMA to optimize the in-memory storage layout and remote transfers between nodes.

\paragraph{\bf PTStore backend for vLLM KV Cache Block Manager (PTStore).}
This is our approach that combines both distributed KV cache prefix awareness
with node-local prefix replication as detailed in \S~\ref{sec:approach}. To the best of our knowledge, this is the first work that exploits both aspects simultaneously.

\subsection{Weak Scalability}
\label{sec:exp:vllm-scalability}

First, we study the weak scalability by using 8, 16, and 32 GPUs and keeping the number of inference queries per GPU constant. As mentioned in \S~\ref{sec:method}, each workers processes 1500 inference requests with a large prompt but small answer, which emphasizes the impact of the prefill phase (initial forward pass) and thus benefits from KV prefix  caching and sharing across different inference queries. For this experiment, we use the wikiQA dataset (~8000 tokens on average per inference request).

\begin{figure*}
    \centering
    \begin{subfigure}{0.49\linewidth}
        \centering
        \includegraphics[width=\linewidth]{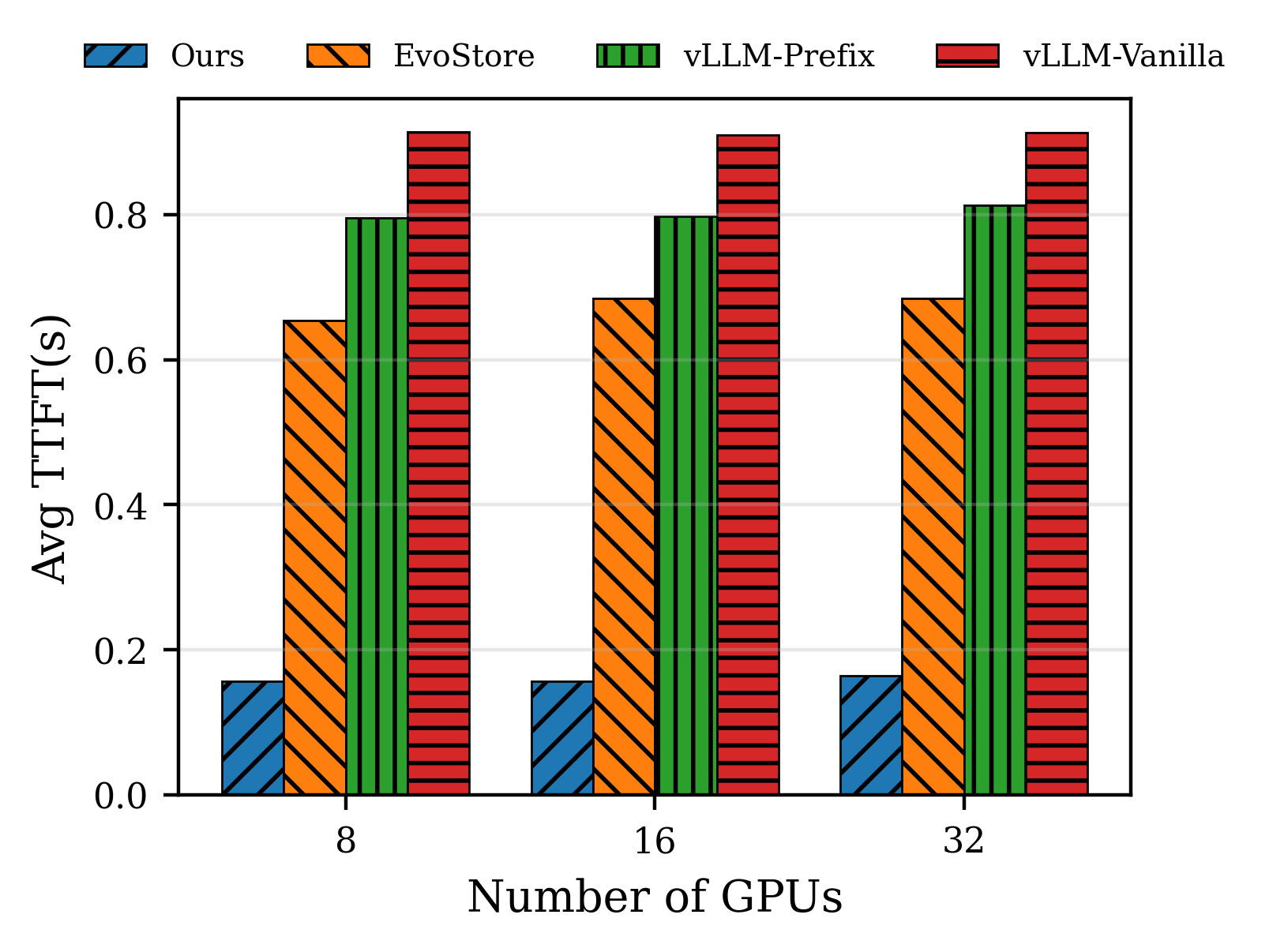}
        \caption{End-to-end comparison between the four approaches.}
         \label{fig:scale:comp}
    \end{subfigure}
    \hfill
    \begin{subfigure}{0.49\linewidth}
        \centering
        \includegraphics[width=\linewidth]{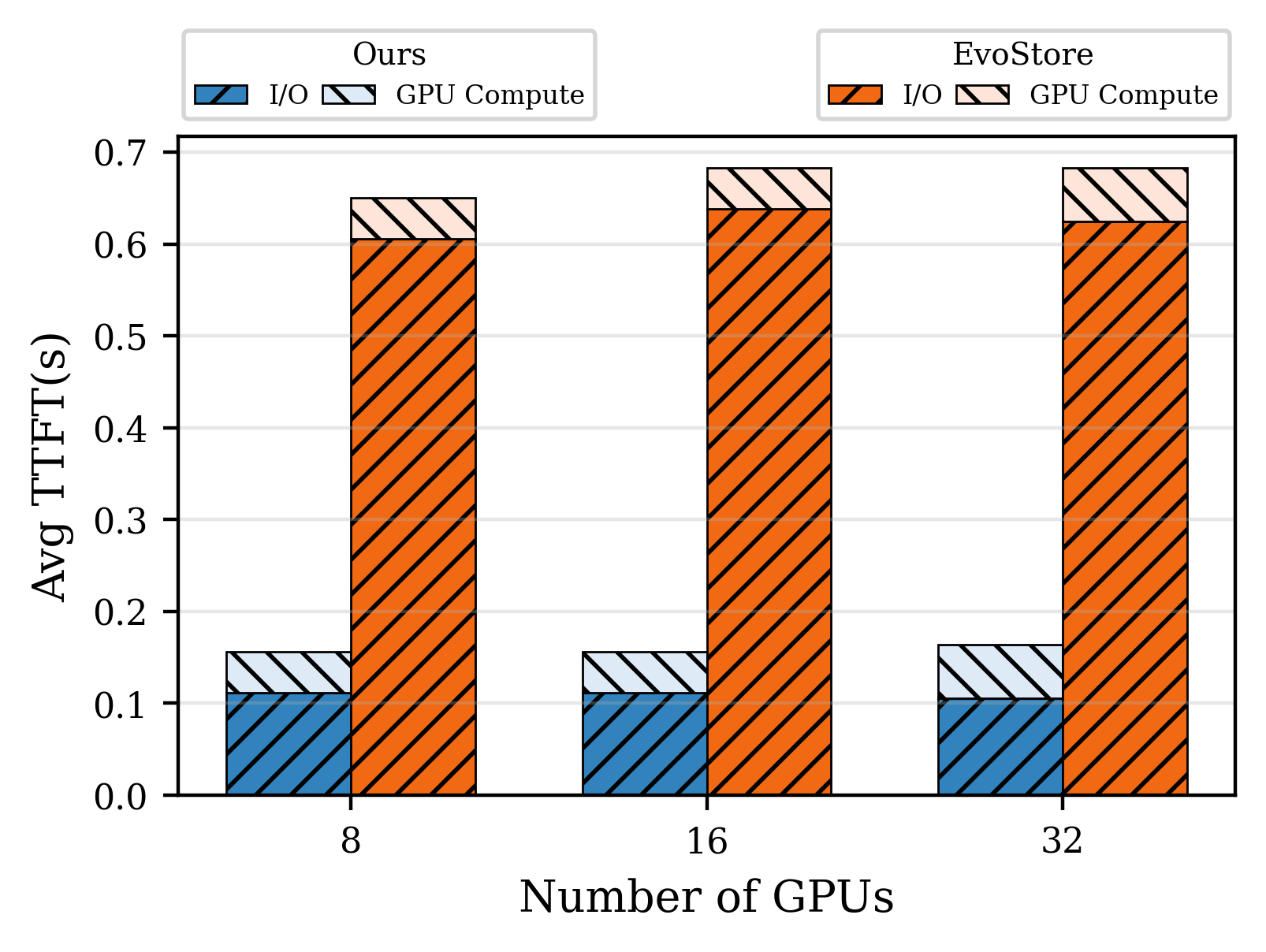}
        \caption{Breakdown of RDMA I/O overheads vs. GPU compute}
         \label{fig:scale:breakdown}
    \end{subfigure}
    \caption{Inference using vLLM: Weak scalability (8-32 GPUs) measured as average time to first token (TTFT). Lower is better.}
    \label{fig:vLLMScale}
\end{figure*}

Figure ~\ref{fig:scale:comp} shows all four approaches scale well, which is particularly important to highlight in the case of \sysname and \emph{EvoStore}, both of which use RDMA for accessing remote memory tiers and therefore are not subject to communication bottlenecks for an increasing scale. However,
the lack of local replication introduces expensive RDMA I/O overheads in the case
of \emph{EvoStore}. This reduces the effectiveness of reusing remote prefixes
to the point where it comes close to simply ignoring remote prefixes, which is the case of \emph{vLLM-Prefix}. In turn, \emph{vLLM-Prefix} is close to
\emph{vLLM-Vanilla}, showing that leveraging only local prefixes misses a lot of
reuse opportunities. Overall, thanks to reusing remote prefixes with low overhead, \sysname has a detached advantage compared with the rest of the approaches.

To further explain the large gap between \sysname and \emph{EvoStore}, we depict
in Figure~\ref{fig:scale:breakdown} a breakdown of the RDMA I/O overheads needed
to fetch remote prefixes vs. the GPU compute overheads needed to run the prefill
phase. As expected, the computational overheads are identical, while the RDMA
I/O overheads are much higher in the case of \emph{EvoStore}, thus confirming
the importance of using local replication to mitigate the negative impact of
these overhead.

\subsection{Sequence Length Scalability}
\label{sec:varyseqlength}

Our next series of experiments studies the effectiveness of prefix caching for an increasing sequence length (sum of input and output of the inference queries). In
this case, we omit vLLM vanilla, since it does not support prefix reuse at all.
We use a variable sequence length ranging from 1000 to 8000 tokens in increments of 2000 tokens (denoted 1k, 2k, 4k, 8k) and measure the average TTFT (time to first token) achieved by the three compared approaches. Unlike the previous experiment, this time we focus on the SQUAD dataset, which has a large number of questions referring to the same text, thus emphasizing the opportunities for prefix reuse and putting more pressure on the caching tiers.

\begin{figure*}
    \centering
    \begin{subfigure}{0.49\linewidth}
        \centering
        \includegraphics[width=\linewidth]{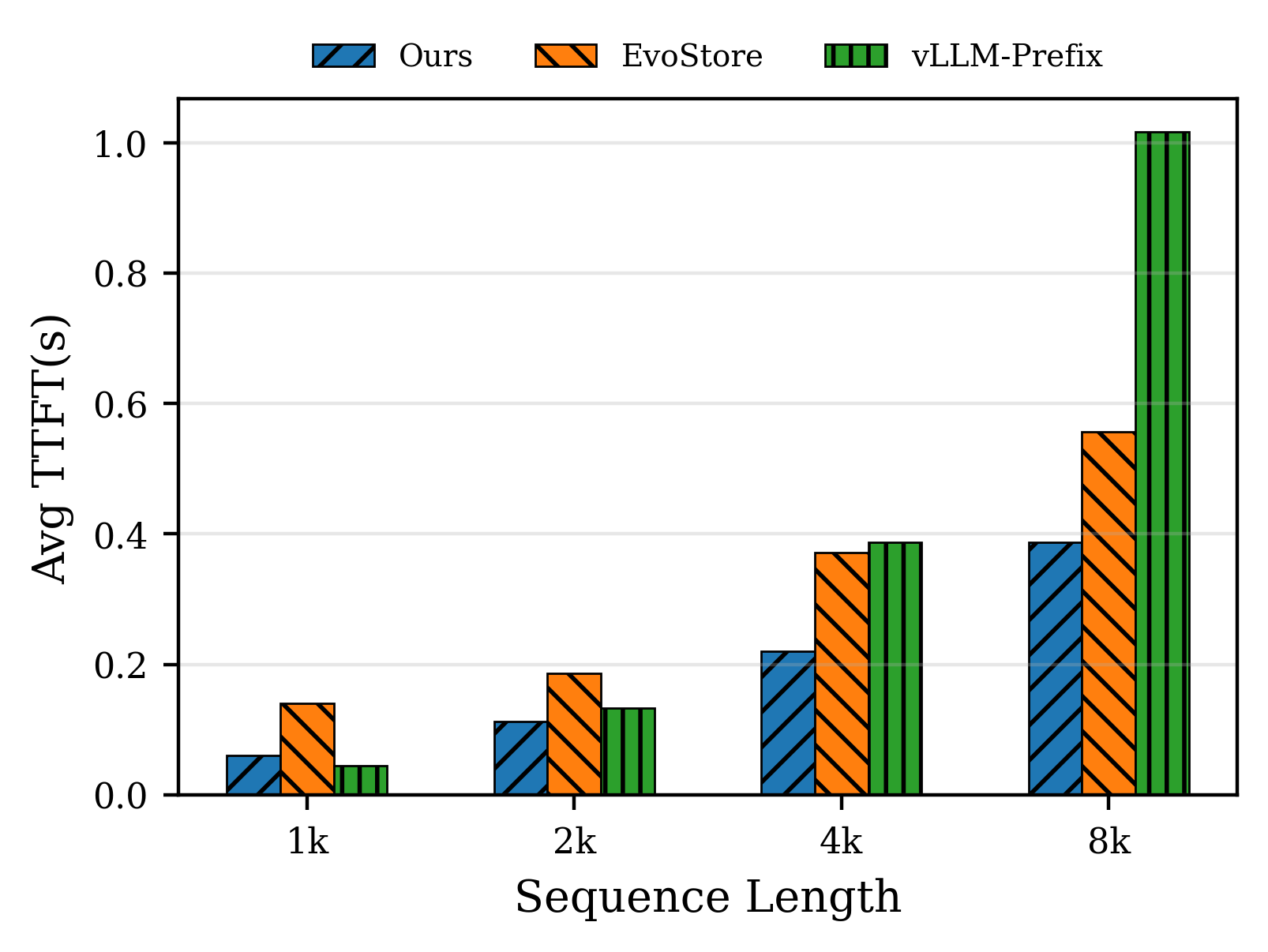}
        \caption{End-to-end comparison between the approaches that support prefix caching.}
         \label{fig:seq:comp}
    \end{subfigure}
    \hfill
    \begin{subfigure}{0.49\linewidth}
        \centering
        \includegraphics[width=\linewidth]{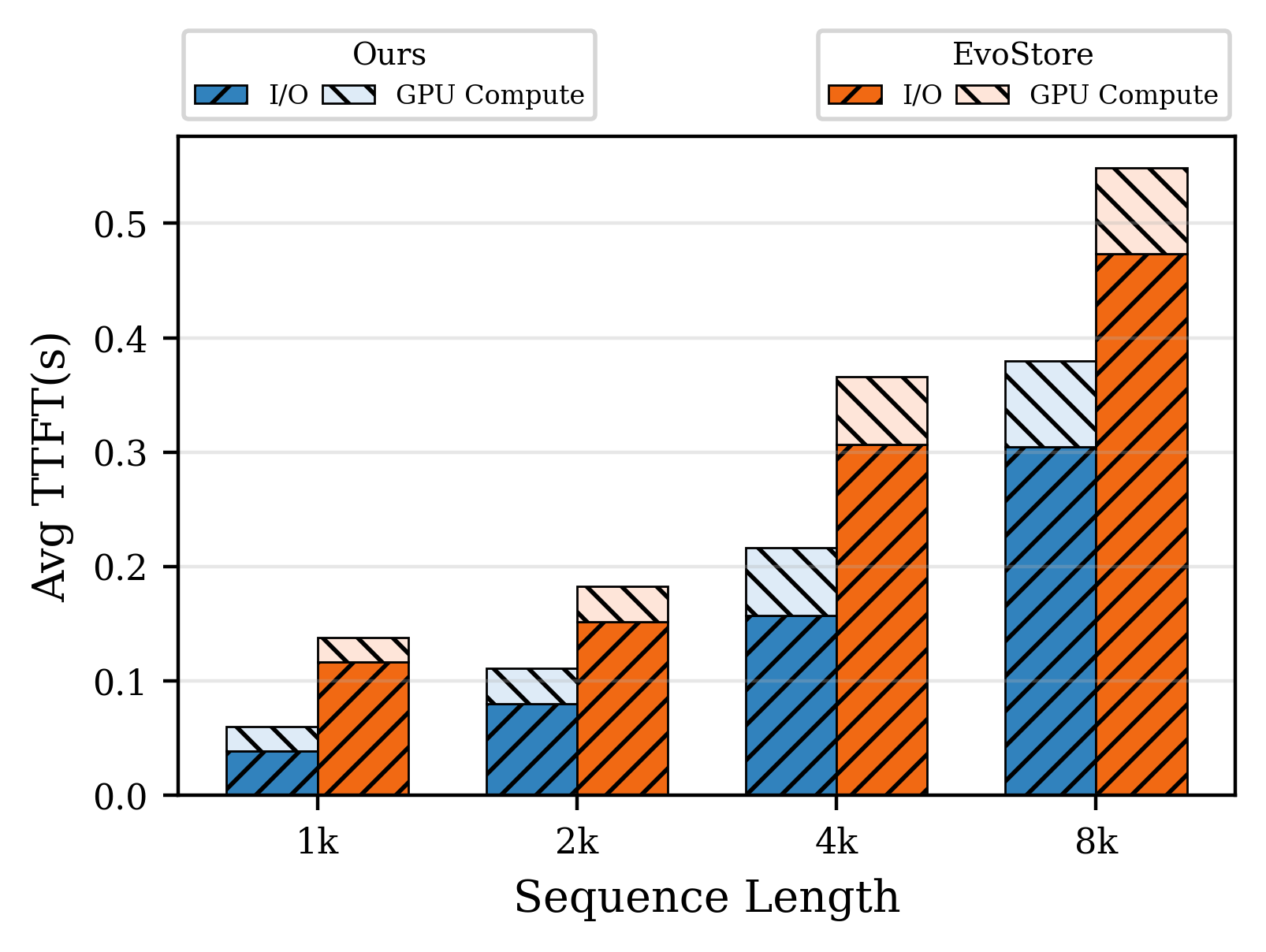}
        \caption{Breakdown of RDMA I/O overheads vs. GPU compute}
         \label{fig:seq:breakdown}
    \end{subfigure}
    \caption{Inference using vLLM: Sequence length scalability (1k-8k tokens) measured as average time to first token (TTFT). Lower is better.}
    \label{fig:vLLMSeq}
\end{figure*}

Figure~\ref{fig:seq:comp} focuses on the end-to-end comparison between the three approaches that support prefix caching. As can be observed, the relative performance improvement of \sysname over vLLM baseline (with support for inter-query prefix caching and reuse activated) increases for an increasing query length. This is expected because at shorter prefix lengths, the corresponding KV cache tensors amount to smaller sizes that can fit in the GPU memory. Also, recomputations are less expensive and offset the benefit of reusing remote prefixes. Thus, for small query lengths, vLLM's baseline prefix caching is superior. However, starting with 2k tokens, the opposite effect is visible: \sysname outperforms vLLM's prefix caching, and the difference between the two approaches keeps growing to the point where \sysname is almost 2x faster for 8k tokens.

Also interesting to note is that the difference between \sysname and \emph{EvoStore}
depends on the sequence length. For 1k tokens, our approach is more than 2x faster.
For 8x tokens, our approach is 20\% faster. As observed in Fig~\ref{fig:seq:breakdown}, a zoom on the breakdown of I/O vs. compute confirms again the advantage of our approach vs. EvoStore thanks to prefix replication.
Furthermore, token distribution also plays an important role, since the results in this case (SQUAD dataset) are not consistent with the weak scalability experiment discussed in \S~\ref{sec:exp:vllm-scalability} (WikiQA dataset).

\section{Conclusions}
\label{sec:conclusions}
This paper introduced \sysname, a scalable distributed repository designed for fine-grained prefix caching and replication to enhance LLM inference throughput. By addressing the limitations of existing systems that lack support for aggregating distributed memory tiers, \sysname enables the efficient reuse of KV cache prefixes across many GPUs and compute nodes. Our design leverages incremental tensor storage to minimize redundancy, consolidated metadata for rapid prefix queries, and an innovative replication strategy that optimizes local access to popular ``hot'' prefixes. Experimental evaluations demonstrate that \sysname significantly reduces Time to First Token (TTFT) vs. state of art, up by an order of magnitude, and
remains scalable for a variable sequence length size. These experiments prove its effectiveness in handling challenging high-throughput inference serving at scale.

Future research will focus on workload-aware mechanisms for dynamic memory balancing, ML-based eviction policies, and ablation studies that isolate the contributions of replication, consolidated metadata, and the replication-cache ratio. We also plan a head-to-head benchmarking of \sysname against LMCache and Mooncake on broader, real-world workloads, including multi-turn conversational agents, code-completion traces, and larger models. Furthermore, we plan tighter integration with diverse inference runtimes and distributed schedulers to further validate \sysname's versatility.

\begin{credits}
\subsubsection{\ackname}
This material is based upon work supported by the U.S. Department of Energy (DOE),  Office of Advanced Scientific Computing Research, under Contract DE-AC02-06CH11357, as well as the National Science Foundation (NSF), Office of Advanced Cyberinfrastructure, CSSI-2411386/241387 and CSSI-2514056.
\end{credits}

\paragraph{\fontsize{9pt}{11pt}\selectfont
The authors have no competing interests to declare relevant to the content of this article.
}

\bibliographystyle{splncs04}
\bibliography{main}

\end{document}